\PassOptionsToPackage{most}{tcolorbox}
\PassOptionsToPackage{table}{xcolor}
\documentclass[11pt]{article}

\usepackage[final]{arxiv}

\usepackage{times}
\usepackage{latexsym}

\usepackage[utf8]{inputenc}
\usepackage{amsmath,amssymb}
\usepackage{pifont}
\usepackage{placeins} 
\usepackage{booktabs}
\usepackage{multirow}
\usepackage{lipsum}
\usepackage{tabularx}
\usepackage{subcaption}
\usepackage{arydshln}
\usepackage{adjustbox} 
\usepackage{makecell}

\usepackage{siunitx}
\usepackage[T1]{fontenc}

\usepackage{inconsolata}
\usepackage{dblfloatfix}
\usepackage{graphicx}
\usepackage{enumitem}
\usepackage{cuted}
\usepackage{capt-of} 
\usepackage{graphicx}
\usepackage{dblfloatfix} 
\usepackage{caption}  
\usepackage{fontawesome5} 
\usepackage{hyperref}     

\usepackage{graphicx}
\usepackage{tikz}
\usetikzlibrary{fadings} 
\usepackage{wrapfig}
\usepackage{svg}
\usepackage{multirow}

\usepackage[utf8]{inputenc}
\usepackage[table]{xcolor} 
\usepackage{array}
\usepackage[most]{tcolorbox}
\usepackage{tcolorbox}
\usepackage{tabularx}
\usepackage{adjustbox}
\usepackage{makecell}

\newtcolorbox{promptbox}[1][]{
    colback=gray!5!white,      
    colframe=gray!75!black,    
    title=\textbf{Prompt Template}, 
    fonttitle=\bfseries\sffamily,
    fontupper=\ttfamily\small, 
    sharp corners, rounded corners=southeast, arc=3mm, 
    boxrule=1pt,               
    enhanced,                  
    breakable,                 
    #1                         
}

\newtcolorbox{keyfinding}[1][]{
  colback=cyan!5!white,    
  colframe=cyan!75!black,  
  fonttitle=\bfseries,     
  title={Key Findings},    
  #1                       
}

\definecolor{leftcolor}{HTML}{0892D0}
\definecolor{rightcolor}{HTML}{4B0082}

\newcommand{\methodname}{\text{TwinBrainVLA}} 

\newcommand{\methodlogo}{\includegraphics[height=\baselineskip]{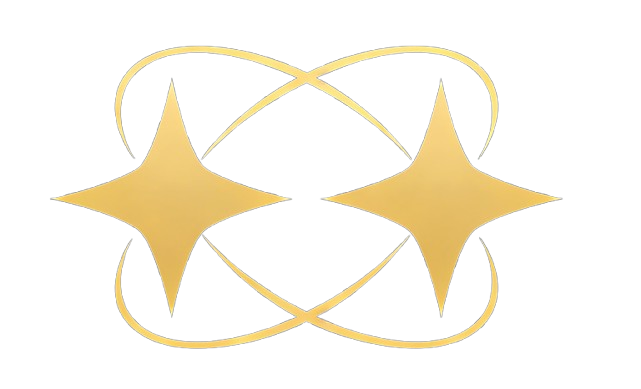}}
\newcommand{\papertitle}{\methodname{}\methodlogo{}: Unleashing the Potential of Generalist VLMs\\for Embodied Tasks via Asymmetric Mixture-of-Transformers}

\title{\papertitle{}}

\author{
  Bin Yu\textsuperscript{1,2,\thanks{Equal contribution}}
  Shijie Lian\textsuperscript{2,4,\footnotemark[1]}
  Xiaopeng Lin\textsuperscript{2,5,\footnotemark[1]}
  Yuliang Wei\textsuperscript{1,\thanks{Corresponding author}}
  Zhaolong Shen\textsuperscript{2,6}
  Changti Wu\textsuperscript{2,7}\\
  \textbf{Yuzhuo Miao\textsuperscript{1,2}}
  \textbf{Xinming Wang\textsuperscript{2,8}}
  \textbf{Bailing Wang\textsuperscript{1}}
  \textbf{Cong Huang\textsuperscript{2,3}}
  \textbf{Kai Chen\textsuperscript{2,3,9,\footnotemark[2]}}
  \\[2ex]
  \textsuperscript{1}HIT\quad
  \textsuperscript{2}ZGCA\quad
  \textsuperscript{3}ZGCI\quad
  \textsuperscript{4}HUST\quad
  \textsuperscript{5}HKUST(GZ)\quad
  \textsuperscript{6}BUAA\quad
  \textsuperscript{7}ECNU\quad
  \textsuperscript{8}CASIA\quad
  \textsuperscript{9}DeepCybo
}

\begin{document}
\maketitle

\begin{center}
    \vspace{-50pt}
    \faGithub\hspace{6pt}\href{https://github.com/ZGC-EmbodyAI/TwinBrainVLA}{\texttt{\color{black}https://github.com/ZGC-EmbodyAI/TwinBrainVLA}}
\end{center}

\vspace{5pt}

\let\thefootnote\relax\footnotetext{Work done at Zhongguancun Academy (Beijing).}

\begin{abstract}

The fundamental premise of Vision-Language-Action (VLA) models is to harness the extensive general capabilities of pre-trained Vision-Language Models (VLMs) for generalized embodied intelligence. However, standard robotic fine-tuning inevitably disrupts the pre-trained feature space, leading to "catastrophic forgetting" that compromises the general visual understanding we aim to leverage. To effectively utilize the uncorrupted general capabilities of VLMs for robotic tasks, we propose \textbf{\methodname{}}\methodlogo{}, which coordinates two isomorphic VLM pathways: a frozen generalist (also called "Left Brain") and a trainable specialist (also called "Right Brain"). Our architecture utilizes a \textbf{Asymmetric Mixture-of-Transformers (AsyMoT)} mechanism, enabling the Right Brain to dynamically query and fuse intact semantic knowledge from the Left Brain with proprioceptive states. This fused representation conditions a flow-matching action expert for precise continuous control. Empirical results on SimplerEnv and RoboCasa benchmarks demonstrate that by explicitly retaining general capabilities, \methodname{} achieves substantial performance gains over baseline models in complex manipulation tasks.
\end{abstract}
\section{Introduction}
\label{sec:intro}

The pursuit of embodied artificial intelligence has recently converged on the paradigm of Vision-Language-Action (VLA) model~\citep{OpenVLA_24,PI05_25,GR00T_25}. By grafting robotic control heads onto pre-trained Vision-Language Models (VLMs), the fundamental premise is to harness the rich semantic reasoning and open-world generalization capabilities learned from internet-scale data to enable generalized robotic control. Ideally, such a system should not merely execute motor commands but leverage its "VLM brain" to understand complex instructions, reason about unseen objects, and plan over long horizons.

However, a critical paradox undermines current VLA architectures: \textbf{the optimization process required to learn low-level sensorimotor control inevitably disrupts the high-level semantic features we aim to leverage.} During standard robotic fine-tuning, the VLM backbone is aggressively updated to minimize action prediction errors on narrow, domain-specific datasets. This creates a conflict between the original objective of semantic understanding and the new objective of proprioceptive control, precipitating "\textbf{\textit{catastrophic forgetting}}"~\citep{VLM2VLA_25,VLM4VLA_25,ChatVLA_25,DualVLA_25}. Consequently, the model sacrifices its pre-trained general capabilities to become a specialized controller. This defeats the original purpose of the VLA paradigm: instead of inheriting a generalized "brain," the fine-tuned model degenerates into a localized policy, stripping away the very open-world understanding required for robust generalization.

Therefore, the central challenge is to effectively utilize the uncorrupted general capabilities of VLMs to enhance robotic performance. We posit that a generalizable VLA model must simultaneously possess both general visual understanding capabilities and low-level servo control proficiency. Drawing inspiration from the biological principle of hemispheric lateralization, where the brain allocates distinct functions to specialized hemispheres, we propose that a VLA should maintain a dedicated "semantic anchor" separate from its "motor controller."



\begin{figure}[h]
    \centering
    \includegraphics[width=0.35\textwidth,clip,trim={20 5 0 5}]{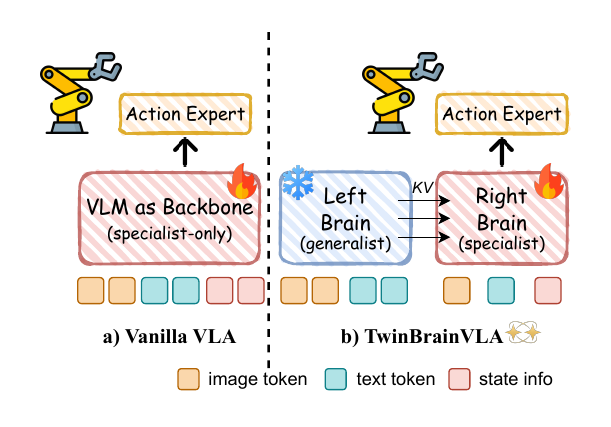}
    \caption{Architectural comparison between Vanilla VLA and \methodname{}.}
    \label{fig:vanilla-vla}
    \vspace{-10pt}
\end{figure}

In this paper, we introduce \textbf{\methodname{}}\methodlogo{}, a novel VLA framework that orchestrates two isomorphic VLM pathways via an asymmetric dual-stream joint training strategy. Structurally, our model comprises a frozen "Left Brain" (generalist) and a trainable "Right Brain" (specialist). The Left Brain preserves the intact, open-world knowledge of the pre-trained VLM, while the Right Brain specializes in processing proprioceptive states and generating actions. Crucially, we propose a novel Asymmetric Mixture-of-Transformers (AsyMoT) mechanism, which enables the trainable Right Brain to dynamically query and fuse semantic features from the frozen Left Brain. This design ensures that the action expert is conditioned on uncorrupted general knowledge, allowing the robot to "think" with a generalist brain while "acting" with a specialist body.

Our contributions are summarized as follows:

\begin{itemize}
    \item We quantitatively analyzed the impairment of the general capabilities of VLMs caused by VLA training through preliminary experiments, namely the occurrence of the "catastrophic forgetting" phenomenon.
    \item We propose \textbf{\methodname{}\methodlogo{}}, an asymmetric dual-stream architecture that structurally decouples semantic understanding from embodied control, allowing the system to leverage pre-trained capabilities without disruption.
    \item We introduce the Asymmetric Mixture-of-Transformers (AsyMoT) mechanism to facilitate efficient information interaction between the two VLM pathways, enabling the specialist Right Brain to dynamically query general knowledge from the frozen Left Brain.
    \item Extensive experiments and evaluations on the SimplerEnv and RoboCasa benchmarks, as well as real-robot settings, demonstrate the effectiveness of the \methodname{} architecture, the AsyMoT mechanism, and our proposed training strategy.
\end{itemize}

\section{Related Work}
\label{sec:related}

\noindent\textbf{Multimodal understanding.}
The landscape of computer vision and natural language processing has been revolutionized by VLMs~\citep{LLaVA_23,Qwen2.5-VL,Qwen3-VL}. By seamlessly integrating visual encoders with Large Language Models (LLMs) through sophisticated projection layers or adapter modules~\citep{Corvid_25}, these models exhibit emergent abilities in semantic understanding and visual question answering (VQA). Given the inherent limitations of general-purpose VLMs in spatial perception~\citep{SpatialTree_25,RynnEC_25,DSIBench_25,SpatialHard_25,CambrianS_25}, recent studies have increasingly focused on employing post-training strategies to enhance spatial intelligence~\citep{VST_25,ReasonRFT_25,RoboRefer_25,LLaVA-3D_25,SpatialForcing_25} and construct specialized embodied foundation models tailored for embodied scenarios~\citep{RoboBrain2_25,VST_25,Mimo-Embodied_25,PhysBrain_25}.

\noindent\textbf{Vision-Language-Action (VLA) Models.} 
Building upon the zero-shot generalization and rich semantic priors of VLMs, VLA models have emerged as a scalable paradigm for embodied intelligence~\cite{FAST_25,UniAct_25,TinyVLA_25,SmolVLA_25,XVLA_25,Evo1_25,InternVLA_A1_25}. By fine-tuning pre-trained VLMs on large-scale robotic manipulation datasets, these models learn to map multimodal instructions and observations to low-level control signals~\citep{G0_25,XR1_25,WallX_25}. To mitigate the degradation of general conversational capabilities in VLMs during VLA training, several approaches have been proposed. LangForce~\citep{LangForce} attempts to enhance VLA performance by introducing mechanisms to mitigate visual shortcuts. ChatVLA~\citep{ChatVLA_25} and VLM2VLA~\citep{ActionAsLanguage_25} attempts to incorporate general dialogue data for co-training; however, this method still suffers from catastrophic forgetting and requires careful curation of general QA data. ThinkAct~\citep{ThinkAct_25} and related works~\citep{DeepThinkVLA_25,OneTwoVLA_25,MolmoAct_25} introduce Chain-of-Thought (CoT) reasoning into VLA training. Nevertheless, current mainstream VLA datasets lack CoT supervision signals, and the annotation process remains expensive and difficult to scale. Instead of relying on extensive data engineering, we propose a structural solution that explicitly decouples semantic preservation from motor learning, enabling the VLA to directly harness the uncorrupted general capabilities of the VLM backbone.
\begin{figure*}[!htb]
    \centering
    \includegraphics[width=1.0\textwidth]{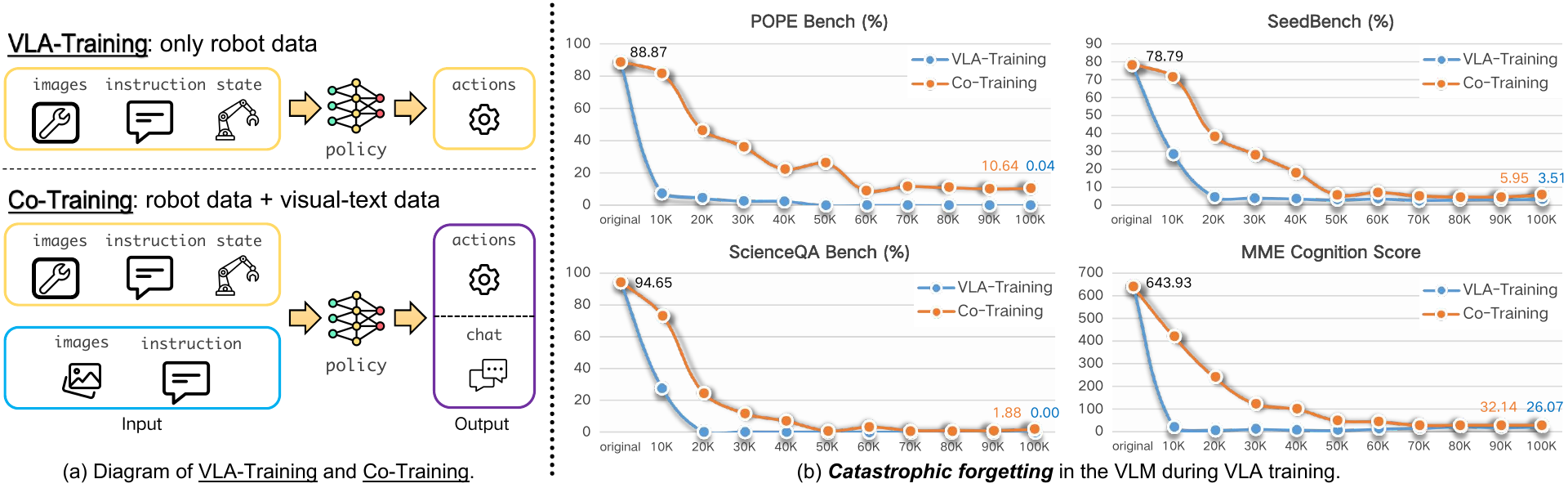}
    \caption{\textbf{Empirical Evidence of Catastrophic Forgetting.} \textbf{(a)} Overview of two prevailing training paradigms: standard VLA-Training (utilizing robot-only data) and Co-Training (mixing robot data with general vision-language data). \textbf{(b)} Evaluation of the VLM’s general visual understanding capabilities across four benchmarks (POPE~\citep{POPE}, SEED-Bench~\citep{SeedBench}, ScienceQA~\citep{ScienceQA}, and MME~\citep{MME}) during the VLA fine-tuning process. Both strategies exhibit a precipitous decline in performance as training steps increase, demonstrating that co-training with general data is insufficient to prevent the collapse of the VLM’s pre-trained semantic reasoning when optimizing for robotic control.}
    \label{fig:motivation}
\end{figure*}

\section{Motivation: The Dilemma of Utilizing General Capabilities in VLA}
\label{sec:motivation}

\subsection{The Conflict Between Specialization and Generalization}
\label{subsec:forgetting}

Standard VLA models are built on a compelling premise: by fine-tuning pre-trained VLMs on robotic datasets, we aim to endow robots with the rich semantic reasoning and open-world understanding inherent in the backbone. Ideally, the VLM should serve as a "generalist brain" that guides the robotic body through complex, unseen environments.

However, a fundamental paradox exists in the current transfer learning paradigm. To adapt the VLM for precise sensorimotor control, the model is fine-tuned on robotic datasets ($\mathcal{D}_{robot}$) that are fundamentally different from the massive internet-scale data used during pre-training. This aggressive optimization for low-level action prediction inevitably disrupts the pre-trained feature space, triggering "\textit{catastrophic forgetting}". This phenomenon turns the VLM backbone from a "generalist" into a "specialist", sacrificing its linguistic brain to gain a robotic body.





\subsection{Empirical Analysis: The Severity of Catastrophic Forgetting}
\label{subsec:empirical_forgetting}

To empirically quantify the extent of catastrophic forgetting discussed above, we conducted a preliminary study using VLM backbones: Qwen2.5-VL-3B-Instruct and Qwen3-VL-4B-Instruct. We evaluate the models' general visual capabilities before and after fine-tuning under two distinct paradigms: \textbf{(1) VLA Training:} The model is fine-tuned exclusively on robotic trajectory data to optimize action generation. \textbf{(2) Co-Training:} Following recent strategies~\citep{HyT_25,ChatVLA_25} to mitigate forgetting, we mix general visual-QA data with robotic data at a 1:1 ratio during fine-tuning, aiming to preserve visual conversational capabilities.

\textbf{Result Analysis.} The results are presented in Figure~\ref{fig:motivation}b. 
First, standard \textbf{VLA Training} leads to a complete collapse of general visual understanding. For instance, the POPE score of Qwen3-VL drops from 88.87\% to near-zero (0.04\%). This confirms that optimizing for proprioceptive action control aggressively overwrites the semantic features required for VQA tasks. More critically, the co-training strategy also fails to prevent this degradation.

\begin{keyfinding}
\textbf{1.} Training VLA models with robotics data induces severe catastrophic forgetting in the VLM, which undermines the original premise of the paradigm that builds VLAs from VLMs. 

\textbf{2.} While Co-Training is frequently suggested as a countermeasure, our experiments show that it merely mitigates the symptoms rather than providing a fundamental solution.
\end{keyfinding}


These findings expose a critical bottleneck: standard fine-tuning effectively erases the 'VLM brain' before it can be deployed for robotic reasoning, rendering it impossible to harness pre-trained general knowledge for complex, open-ended tasks. While the success of multimodal applications has proven that general understanding is essential for downstream performance which is also a key insight driving the VLA paradigm. However, current training paradigms lead to severe catastrophic forgetting, which directly contradicts the initial design philosophy of VLAs. Therefore, the challenge remains: \textit{\textbf{How can we construct VLA models to leverage the uncorrupted multimodal understanding capabilities of VLMs to enhance their own performance?}}


\section{Method: \methodname{}\methodlogo{}}
\label{sec:method}

In this section, we present \textbf{\methodname{}\methodlogo{}}, a novel framework designed to enable VLA models to effectively utilize the general visual understanding capabilities of pre-trained VLMs for embodied tasks. While standard monolithic architectures suffer from feature degradation during fine-tuning, our approach aims to retain the full spectrum of pre-trained semantic knowledge and make it explicitly accessible for robotic control. To this end, we introduce an asymmetric dual-stream design that structurally separates the preservation of general capabilities from the learning of sensorimotor skills, ensuring that the control policy can leverage uncorrupted multimodal understanding capabilities.


\begin{figure}[htbp]
    \centering
    \includegraphics[width=1.0\textwidth]{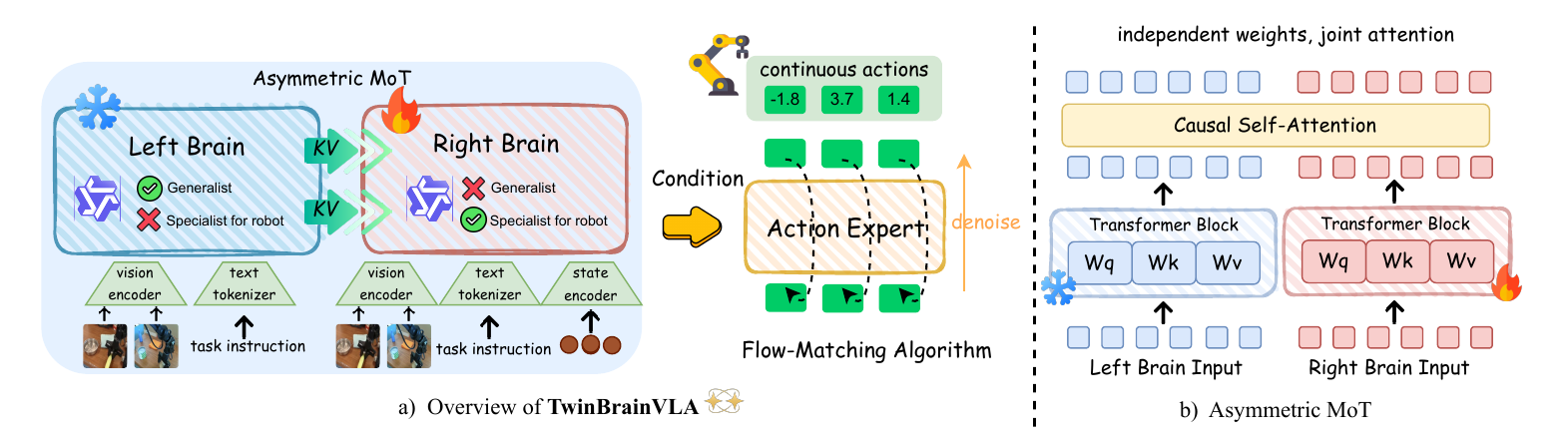}
    \caption{\textbf{The framework of \methodname{}\methodlogo{}.} \textbf{(a) Overall Architecture.} The model features an Asymmetric Mixture-of-Transformers design composed of two distinct pathways: a frozen "Left Brain" (Generalist) for semantic reasoning and a trainable "Right Brain" (Specialist) for embodied motor control. The Right Brain fuses visual, textual, and proprioceptive state inputs to provide conditioning for the Action Expert, which utilizes a Flow-Matching algorithm to denoise continuous robotic actions. \textbf{(b) Asymmetric MoT Mechanism (AsyMoT).} Through causal self-attention, the trainable Right Brain attends to the frozen Key-Value (KV) pairs of the Left Brain, enabling the transfer of general semantic knowledge to the robotic control policy without catastrophic forgetting.}
    \label{fig:arch}
\end{figure}

\subsection{Asymmetric Dual-VLM Backbone}
\label{subsec:dual_vlm}

The backbone of \methodname{} comprises two parallel VLM pathways: the "Left Brain" ($\mathcal{M}_{L}$) and the "Right Brain" ($\mathcal{M}_{R}$). Both streams are initialized with identical pre-trained weights (e.g., Qwen3-VL series) but play distinct roles to facilitate the utilization of general capabilities for robot control.

\textbf{The "Left Brain" as a General Capability Reservoir.} The Left Brain serves as the provider of general visual understanding. To ensure that the pre-trained capabilities remain intact, it is kept strictly frozen throughout the training phase. It processes the visual observation $I$ and instruction $T$ to generate semantic representations:

\begin{equation}
    H_{L}^{0}=[\mathcal{V}(I);\mathcal{T}(T)]
\end{equation}

where $\mathcal{V}$ and $\mathcal{T}$ denote the vision encoder and text tokenizer. By freezing this pathway, the Left Brain acts as a stable semantic anchor, offering a continuous source of open-world knowledge that is immune to the catastrophic forgetting typically induced by robotic fine-tuning.

\textbf{The "Right Brain" as a Semantic-Augmented Controller.} The Right Brain is a trainable specialist responsible for executing robotic actions. To ground the model in the physical world, we introduce a lightweight State Encoder $\phi$, which projects the robot's proprioceptive state $s$ (e.g., joint angles) into the VLM's embedding space:

\begin{equation}
    H_{R}^{0}=[\mathcal{V}(I);\mathcal{T}(T);\phi(s)]
\end{equation}

Unlike a standard VLA that relies solely on its own degrading parameters, the Right Brain is designed to actively harness the capabilities of the Left Brain to inform its decision-making.



\noindent\textbf{Asymmetric Mixture-of-Transformers (AsyMoT).}
Mixture-of-Transformers (MoT)~\citep{MoT} is a model architecture that implements forward propagation by connecting two independent Transformer networks via joint attention at each corresponding layer. Its advantage lies in the ability to facilitate information interaction between multiple modalities while still maintaining their independent computation. Inspired by this work, we propose the Asymmetric Mixture-of-Transformers (AsyMoT) mechanism to enable the Right Brain to utilize the general visual understanding from the Left Brain. This mechanism establishes a unidirectional information bridge, allowing the control policy to query the frozen general knowledge.

Let $\mathbf{H}_L^l$ and $\mathbf{H}_R^l$ denote the hidden states of the Left and Right Brains at layer $l$, respectively.

the Left Brain remains \textit{frozen} during training. Its self-attention mechanism operates independently to preserve pre-trained general capabilities:
\begin{equation}
    \mathbf{H}_L^{l+1} = \text{Attn}(Q_L^l, K_L^l, V_L^l) + \text{FFN}(\mathbf{H}_L^l),
\end{equation}
where $Q, K, V$ are derived from $\mathbf{H}_L^l$ using frozen projection weights.

The Right Brain is \textit{trainable} and employs an asymmetric joint attention mechanism. It queries not only its own context but also the semantic features from the Left Brain. Specifically, at each attention head, the Right Brain computes its Query ($Q_R$), while the Key ($K$) and Value ($V$) are constructed by concatenating the representations from both brains:
\begin{align}
    K_{\text{joint}} &= [\text{sg}(K_L^l) \ ; \ K_R^l], \\
    V_{\text{joint}} &= [\text{sg}(V_L^l) \ ; \ V_R^l], \\
    \mathbf{H}_R^{l+1} &= \text{Softmax}\left(\frac{Q_R^l (K_{\text{joint}})^T}{\sqrt{d_k}}\right) V_{\text{joint}} + \text{FFN}(\mathbf{H}_R^l),
\end{align}
where $[\ ; \ ]$ denotes concatenation along the sequence length dimension, and $\text{sg}(\cdot)$ indicates the stop-gradient operation. 


\textbf{AsyMoT v.s. Cross-Attention.} The key to the AsyMoT mechanism lies in the asymmetric joint attention at each layer. It allows the 'right brain' to attend to both the 'left brain' and itself, whereas the 'left brain' is restricted to attending only to itself, resulting in an asymmetric phenomenon. In contrast, Cross Attention is a different multimodal fusion architecture. It enables a Query from one modality to attend to the Key-Values of another modality, but precludes it from attending to itself. This constitutes the critical distinction between AsyMoT and Cross Attention."

\subsection{Flow-Matching Action Expert}
\label{subsec:action_expert}

Following the paradigm of recent mainstream VLAs~\citep{GR00T_25, PI05_25}, we employ a generative action expert based on flow matching to enable high-precision continuous control. 
This module is implemented as a Diffusion Transformer (DiT)~\citep{DiT_23}, which operates as a conditional decoder.
Specifically, it takes the representations $\mathbf{H}_R$ from the Right Brain as the condition to synthesize continuous action trajectories from noise.
The training objective is defined by the standard vector field regression loss:
\begin{equation}
    \mathcal{L}_{\text{action}}(\psi) = \mathbb{E}_{t, \mathbf{a}_0, \mathbf{a}_1} \left[ || v_\psi(\mathbf{a}_t, t, \mathbf{H}_R) - (\mathbf{a}_1 - \mathbf{a}_0) ||^2 \right],
\end{equation}
where $v_\psi$ denotes the DiT network, $\mathbf{a}_0$ is the Gaussian noise, and $\mathbf{a}_1$ represents the ground-truth action.

\subsection{Training Strategy}
\label{subsec:training_strategy}

\noindent\textbf{Optimization Objective.}
Consistent with standard VLA fine-tuning paradigms \citep{GR00T_25}, we train \methodname{} using exclusively the robotic action objective. The training objective is to minimize the Flow-Matching loss:

\begin{equation}
    \mathcal{L}_{\text{total}} = \mathcal{L}_{\text{action}}(\theta_R, \psi, \phi; \mathcal{D}_{\text{robot}}),
\end{equation}

where $\mathcal{D}_{\text{robot}}$ represents the robotic demonstration dataset, and $\{\theta_R, \psi, \phi\}$ denote the trainable parameters of the Right Brain, Action Expert, and State Encoder, respectively. 
It is worth noting that in monolithic VLA architectures, relying solely on action loss typically leads to severe catastrophic forgetting of general semantic capabilities. However, our dual-stream design structurally immunizes the model against this degradation: the "Right Brain" is free to specialize entirely in control dynamics, while the frozen "Left Brain" implicitly safeguards the linguistic and semantic priors.


\section{Experiment}
\label{sec:experiment}

To comprehensively evaluate the efficacy of \methodname{}, we conduct extensive experiments on two simulation benchmarks: \textbf{SimplerEnv}~\citep{SimplerEnv_24} and \textbf{RoboCasa}~\citep{RoboCasa_24}. Our training pipeline is built upon the starVLA framework~\citep{starvla_2025}, distributed across 16 $\times$ NVIDIA H100 GPUs. We strictly follow its default training protocols to ensure a fair comparison. More extensive simulation benchmarks and real-world robotic experiments are in progress.

\subsection{Experiments on SimplerEnv (OOD)}
\label{subsec:experiment_on_simplerenv}

\definecolor{navyblue}{HTML}{0071BC}

\begin{table*}[ht]
  \centering
  \caption{
    \textbf{Results of evaluating the VLA models with the WidowX robot in the SimplerEnv simulation environment}. We highlight the best results in \textbf{bold} and the second-best results with \underline{underline}. Here, \textbf{Vanilla VLA} refers to a VLA constructed with a single VLM backbone. Its only difference from TwinBrainVLA is the removal of the frozen Left Brain.
    }
  \begin{adjustbox}{width=\linewidth}
  
  \begin{tabular}{l c c c c c}
    \toprule
    \textbf{Method}
     & \makecell[c]{\textbf{Put Spoon} \\ \textbf{on Towel}} 
     & \makecell[c]{\textbf{Put Carrot} \\ \textbf{on Plate}} 
     & \makecell[c]{\textbf{Stack Green Block} \\ \textbf{on Yellow Block}} 
     & \makecell[c]{\textbf{Put Eggplant} \\ \textbf{in Yellow Basket}} 
     & \textbf{Average} \\
    \midrule

    RT-1-X~\citep{OXE_24}         &  0.0  & 4.2   & 0.0   & 0.0   & 1.1 \\
    Octo-Base~\citep{Octo_2024}       & 15.8  & 12.5  & 0.0   & 41.7  & 17.5 \\
    Octo-Small~\citep{Octo_2024}      & 41.7  & 8.2   & 0.0   & 56.7  & 26.7 \\
    OpenVLA~\citep{OpenVLA_24}         & 4.2   & 0.0   & 0.0   & 12.5   & 4.2 \\
    RoboVLM~\citep{RoboVLM_2024}         & 50.0  & 37.5  & 0.0   & 83.3  & 42.7 \\ 
    TraceVLA~\citep{TraceVLA_2025}        & 12.5  & 16.6  & 16.6  & 65.0  & 27.7 \\
    SpatialVLA~\citep{SpatialVLA_2025}      & 20.8  & 20.8  & 25.0  & 70.8  & 34.4 \\
    ThinkAct~\citep{ThinkAct_25} & 58.3 & 37.5 & 8.7 & 70.8 \\
    CogACT~\citep{CogACT_2024}          & 71.7 &  50.8  & 15.0 & 67.5 & 51.3 \\
    VideoVLA~\citep{VideoVLA_2025}        & 75.0 & 20.8   & 45.8 & 70.8 & 53.1 \\
    $\pi_0$~\citep{PI0}         & 29.1 & 0.0 & 16.6 & 62.5 & 27.1 \\
    $\pi_{0.5}$~\citep{PI05_25} & 49.3 & 64.7 & 44.7 & 69.7 & \underline{57.1} \\


    Isaac-GR00T-N1.6-Bridge~\citep{GR00T_N1.6}   & 64.5 & 65.5 & 5.5 & 93.0 & \underline{57.1} \\
    
    \cmidrule(lr){1-6}  

    \textbf{Vanilla VLA} + Qwen2.5-VL-3B-Instruct  & 74.9 & 44.7 & 9.3 & 81.2 & 52.5 \\
    \textbf{Vanilla VLA} + Qwen2.5-VL-7B-Instruct & 68.7 & 35.4 & 16.7 & 62.5 & 44.8 \\
    \textbf{Vanilla VLA} + Qwen3-VL-4B-Instruct  & 87.5 &  50.0 & 29.2 & 54.2 & 55.2 \\
    \textbf{Vanilla VLA} + Qwen3-VL-8B-Instruct  & 68.7 & 38.5 & 30.2 & 87.9 & 56.3 \\

    \midrule
    \rowcolor{navyblue!10}\multicolumn{6}{c}{{\textit{\textbf{ours}}}} \\

    \textbf{\methodname{}} + Qwen2.5-VL-3B-Instruct  &  83.3 &  41.7 & 31.5 & 77.1 & \textbf{58.4} \\
    \textbf{\methodname{}} + Qwen3-VL-4B-Instruct  &  87.5 &  58.3 & 33.3 & 79.1 & \textbf{64.5} \\
    
    \bottomrule
  \end{tabular}
  \end{adjustbox}
  \vspace{0.5 em}
  \label{tab:simplerenv_main_tab}
\end{table*}

\noindent\textbf{Implementation and Training Setup.}
To demonstrate the scalability and effectiveness of our architecture, we instantiate \methodname{} with two state-of-the-art VLM backbones: Qwen2.5-VL-3B-Instruct and Qwen3-VL-4B-Instruct. Consistent with our asymmetric design, both the frozen Left Brain and the trainable Right Brain are initialized from these pre-trained checkpoints to ensure aligned feature spaces. For the training data, we utilize two large-scale subsets from the Open X-Embodiment (OXE) dataset: the \texttt{Bridge-V2} and the \texttt{Fractal} dataset. Comprehensive implementation details are provided in Appendix~\ref{app:hyperparameters}.

\noindent\textbf{Evaluation.}
The benchmark consists of four manipulation tasks: "put spoon on towel", "put carrot on plate", "stack green block on yellow block", "put eggplant in the yellow basket". For each task, we evaluate our VLA policy using the evaluation script provided by the SimplerEnv repository~\citep{SimplerEnv_24}. To mitigate randomness, we run 480 independent trials and report the mean performance (Avg@480). Since the evaluation environments are completely absent from the training set, this constitutes an Out-of-Domain (OOD) evaluation.


\noindent\textbf{Results.}
The quantitative results on the SimplerEnv benchmark are presented in Table~\ref{tab:simplerenv_main_tab}. Notably, despite not undergoing large-scale pre-training for robotic action prediction, \methodname{} achieves state-of-the-art performance among all listed methods. Our framework demonstrates strong generalizability across different VLM families, attaining competitive success rates of 58.4\% with Qwen2.5-VL-3B-Instruct and \textbf{64.5\%} with Qwen3-VL-4B-Instruct. The latter configuration surpasses the strongest baseline, Isaac-GR00T-N1.6 (57.1\%), by a notable margin of +7.4\%, validating the effectiveness of our asymmetric dual-brain architecture in bridging high-level semantic understanding and low-level robotic control.

\subsection{Experiments on RoboCasa}
\label{subsec:experiment_on_robocasa}

\noindent\textbf{Implementation.}
We train \methodname{} on the Humanoid Robot Tabletop Manipulation subset from the PhysicalAI-Robotics-GR00T-X-Embodiment-Sim dataset~\citep{GR00T_25}. All other experimental settings, including model initialization, distributed training infrastructure, and hyperparameters, remain identical to those described in Sec.~\ref{subsec:experiment_on_simplerenv}.

\begin{table}[!t]
    \centering
    \small

    \caption{
      \textbf{Results of evaluating the VLA models with the GR1 robot in the RoboCasa Tabletop simulation environment}. The results for Isaac-GR00T N1.5 and Isaac-GR00T N1.6 are sourced from the official Isaac-GR00T github repository~\citep{GR00T_N1.6}. The results for the first four baseline methods are sourced from the official starVLA experiments~\citep{starvla_2025}. Performance on all 24 tasks can be found in Table~\ref{tab:robocasa_main_tab} of the Appendix.
    }
    \begin{adjustbox}{width=0.75\columnwidth}
    \begin{tabular}{l c c c c}
        \toprule
        \rowcolor{white} 
        {Model} & 
        {\scriptsize \makecell{QwenFAST\\+Qwen3VL}} &
        {\scriptsize \makecell{QwenGR00T\\+Qwen3VL}} & 
        {\scriptsize \makecell{QwenPI\\+Qwen3VL}} & 
        {\scriptsize \makecell{QwenOFT\\+Qwen3VL}} \\
        \midrule
        \textbf{Average} & 39.0 & 47.8 & 43.9 & 48.8 \\
        \midrule
        \rowcolor{white} 
        {Model} & 
        {\scriptsize \makecell{Isaac-GR00T\\N1.5}} & 
        {\scriptsize \makecell{Isaac-GR00T\\N1.6}} & 
        {\scriptsize \makecell{\textbf{\methodname} \\ + Qwen2.5VL}} & 
        {\scriptsize \makecell{\textbf{\methodname} \\ + Qwen3VL}} \\
        \midrule
        \textbf{Average} & 48.2 & 47.6 & \underline{53.5} & \textbf{54.6} \\
        \bottomrule
    \end{tabular}
    \end{adjustbox}
    \label{tab:robocasa_brief_tab}
\end{table}

\noindent\textbf{Evaluation.}
We evaluate our policy on the RoboCasa GR1 Tabletop Benchmark, which comprises a diverse suite of 24 tabletop manipulation tasks. These tasks involve complex interactions with articulated objects and varying geometries, such as \texttt{PnPBottleToCabinetClose} (placing a bottle into a cabinet and closing the door) and \texttt{PnPCanToDrawerClose}.
To ensure statistical reliability while maintaining consistency with our previous experiments, we report the success rate averaged over 50 independent trials for each of the 24 tasks.


\noindent\textbf{Results.}
The quantitative results on the RoboCasa GR1 Tabletop Benchmark are presented in Table~\ref{tab:robocasa_main_tab}. \methodname{} achieves the best performance across all 24 manipulation tasks, with our Qwen3-VL-4B-Instruct variant attaining an average success rate of \textbf{54.6\%}, followed closely by the Qwen2.5-VL-3B-Instruct variant at 53.5\%. These results substantially outperform all baseline methods: our best model surpasses Isaac-GR00T-N1.6 (47.6\%) by +7.0\%, QwenGR00T (47.8\%) by +6.8\%, and QwenPI (43.9\%) by +10.7\%. These results reinforce our hypothesis that decoupling semantic understanding from embodied perception enables more effective learning of fine-grained manipulation skills in complex tabletop scenarios.

\subsection{Experiments on LIBERO Benchmark}

We also evaluate \methodname{} on the LIBERO benchmark~\citep{libero}. However, given that current VLA research has saturated this benchmark (over 95\%) and its tasks are strictly in-domain, it offers limited insight into the benefits of general semantic preservation. To better evaluate the model's generalization capabilities, we combined the training data from all four LIBERO task suites for VLA training and evaluated a single model across all four suites. In this setting, \methodname{} attains a 97.6\% average success rate without relying on massive pre-training. See Appendix~\ref{app:libero} for implementation and evaluation details.

\begin{table}[h]
    \centering
    \small
    \caption{\textbf{Comparison on the LIBERO benchmark.}}
    \label{tab:libero_results}
    \setlength{\tabcolsep}{4pt} 
    \begin{tabular}{lccccc}
        \toprule
        \textbf{Method} & \textbf{Spatial} & \textbf{Object} & \textbf{Goal} & \textbf{Long} & \textbf{Avg} \\
        \midrule
        OpenVLA & 87.4 & 88.4 & 79.2 & 53.7 & 76.5 \\
        $\pi_0$ & 96.8 & 98.8 & 95.8 & 85.2 & 94.1 \\
        $\pi_{0.5}$ & 98.8 & 98.2 & 98.0 & 92.4 & 96.9 \\
        \rowcolor{gray!20}
        \textbf{\methodname{}} & 99.2 & 99.0 & 96.8 & 95.4 & \textbf{97.6} \\
        \bottomrule
    \end{tabular}
\end{table}

\subsection{Real-robot Experiments}

To demonstrate the practical applicability of \methodname{} in real world, we conduct experiments using a Franka Research 3 (7-DOF).

\begin{figure}[h]
    \centering
    \includegraphics[width=0.45\textwidth,clip,trim={1 1 1 1}]{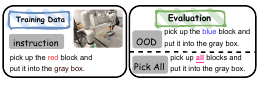}
    \caption{Real-robot Experiment Environment.}
    \label{fig:franka}
    \vspace{-10pt}
\end{figure}

\noindent\textbf{Pre-training Setup.}
We curate a high-quality pre-training corpus by selecting 20 subsets from the Open X-Embodiment (OXE) dataset. Our selection criteria prioritize high video resolution, diverse task instructions, and unified end-effector position control formats. This dataset covers multiple robot embodiments, including WidowX, Google Robot, and Franka (see Appendix~\ref{app:real-robot} for details). \methodname{} is pre-trained on this datasets to acquire generalizable sensorimotor priors before specific task adaptation.

\noindent\textbf{Fine-tuning and Evaluation.}
For downstream adaptation, we collect 300 teleoperated demonstration trajectories on the Franka robot for a pick-and-place task (instruction: \textit{``pick up the red block and put it into the gray box''}). 
To rigorously assess robustness, we design three evaluation protocols: 
(1) \textbf{In-Domain:} The test environment and objects are identical to the training data; 
(2) \textbf{Out-of-Domain (OOD):} The target object's attributes (e.g., block color) are altered to evaluate zero-shot visual generalization.
(3) \textbf{Pick-All:} In this task, the robot is required to place all blocks into the box. We expect the model to complete this long-horizon task by leveraging the atomic actions learned from the training set.
We report the average success rate over 30 independent trials for each setting.

\textbf{Results.} In real-robot experiments, despite utilizing significantly less pre-training data than $\pi_{0.5}$, our method achieves comparable performance after fine-tuning on teleoperation data. Furthermore, it demonstrates robust results in both OOD and Pick-All evaluations. This validates the effectiveness of our model architecture and underscores the importance of retaining generalizability for task generalization.

\begin{table}[h]
    \centering
    \scriptsize
    \caption{\textbf{Real-Robot Experiment Results.} Success rates (\%) of the pick-and-place task on the Franka Research 3 robot.}
    \label{tab:real_robot}
    \setlength{\tabcolsep}{6pt} 
    \begin{tabular}{lccc}
        \toprule
        \multirow{2}{*}{\textbf{Method}} & \multicolumn{3}{c}{\textbf{Success Rate}} \\
        \cmidrule(lr){2-4}
        & \textbf{In-Domain} & \textbf{Out-of-Domain} & \textbf{Pick-All} \\
        \midrule
        OpenVLA~\citep{OpenVLA_24} & 7/30 & 0/30 & 0/30 \\
        $\pi_0$~\citep{PI0} & 23/30 & 9/30 & 0/30 \\
        $\pi_{0.5}$~\citep{PI05_25} & \textbf{28/30} & \underline{13/30} & \underline{2/30} \\
        Vanilla VLA & 22/30 & 9/30 & 0/30 \\
        \rowcolor{gray!20}
        \textbf{\methodname{} (Ours)} & \textbf{28/30} & \textbf{15/30} & \textbf{3/30} \\
        \bottomrule
    \end{tabular}
    \vspace{-1em}
\end{table}

\subsection{Ablation Experiments}

\textbf{Impact of asymmetric freezing strategy.}
To validate our asymmetric design, we ablate the freezing strategy by making the Left Brain trainable. Results show that this configuration hurts performance, causing a \textbf{7\% drop} in success rate for the Qwen3-VL-4B backbone on SimplerEnv. This result highlights that explicitly retaining general capabilities as an accessible resource is paramount for achieving high performance in VLA tasks.

\definecolor{navyblue}{HTML}{0071BC}

\begin{table}[h]
  \centering
  \small
  \caption{\textbf{Ablation on Freezing Strategy.} Quantitative evaluation of success rates (\%) on SimplerEnv tasks. We benchmark \methodname{} against the Single-Stream VLA (QwenGR00T) under identical training settings using Qwen2.5-VL-3B and Qwen3-VL-4B backbones.}
  \begin{adjustbox}{width=0.5\columnwidth}
  
  \begin{tabular}{l c c c c c}
    \toprule
    \textbf{Method}
     & \makecell[c]{\textbf{Put Spoon} \\ \textbf{on Towel}} 
     & \makecell[c]{\textbf{Put Carrot} \\ \textbf{on Plate}} 
     & \makecell[c]{\textbf{Stack} \\ \textbf{Block}} 
     & \makecell[c]{\textbf{Eggplant} \\ \textbf{in Basket}} 
     & \textbf{Avg} \\
    \midrule
    \rowcolor{navyblue!10}\multicolumn{6}{c}{{\textit{\textbf{Qwen2.5-VL-3B-Instruct}}}} \\
    No Freezing  & 64.6 & 45.8 & 20.8 & 64.6 & 49.0 \\
    \textbf{\methodname{}}  &  83.3 &  41.7 & 31.5 & 77.1 & \textbf{58.4} \\
    \rowcolor{navyblue!10}\multicolumn{6}{c}{{\textit{\textbf{Qwen3-VL-4B-Instruct}}}} \\
    No Freezing  & 85.4 &  50.0 & 29.1 & 70.8 & 58.8 \\
    \textbf{\methodname{}} &  87.5 &  58.3 & 33.3 & 79.1 & \textbf{64.5} \\
    \bottomrule
  \end{tabular}
  \end{adjustbox}
  \label{tab:abla_single_stream}
\end{table}

\textbf{Impact of interaction density in AsyMoT.}
In TwinBrainVLA, joint attention is performed layer-wise by default. We further investigated the impact of interaction density by sparsifying the connections. Specifically, we experimented with interaction intervals of $k=\{0, 1, 2, 4, 8\}$ layers. When $k=1$, the asymmetric joint attention and naive self-attention are performed alternately. The results in Table~\ref{tab:ablation_asymot} demonstrate that effective information fusion yields performance gains, whereas performance degrades when the connections become overly sparse.

\textbf{Impact of dual-VLM.} If the left brain and the attention connections between the two VLMs are removed, the network architecture degenerates into a Vanilla VLA. As shown in Table~\ref{tab:simplerenv_main_tab}, this degradation leads to a performance drop of nearly 7 percentage points.

\definecolor{navyblue}{HTML}{0071BC}

\begin{table}[!t]
    \centering
    \small
    \caption{\textbf{Ablation on AsyMoT Interaction Frequency.} We investigate the impact of sparsifying the joint attention connections by varying the interaction interval $k$ on SimplerEnv benchmark.}
    \label{tab:ablation_asymot}
    \begin{adjustbox}{width=0.7\columnwidth}
    \begin{tabular}{lccccc}
        \toprule
        \textbf{Interval ($k$)} & \textbf{0} (ours) & \textbf{1} & \textbf{2} & \textbf{4} & \textbf{8} \\
        \midrule
        Qwen2.5-VL-3B-Instruct based & \underline{58.4} & \textbf{59.4} & 57.4 & 56.5 & 55.9 \\
        Qwen3-VL-4B-Instruct based & \underline{64.5} & 64.1 & \textbf{66.7} & 62.5 & 61.9 \\
        \bottomrule
    \end{tabular}
    \end{adjustbox}
\end{table}
\section{Discussion}
\label{sec:discussion}


\subsection{From Forgetting Mitigation to Capability Utilization}
\methodname{} effectively prevents catastrophic forgetting by maintaining a strictly frozen "Left Brain" throughout the training process. This strategy explicitly preserves the general semantic understanding capabilities acquired during VLM pre-training, ensuring that the model retains its robust open-world knowledge while the "Right Brain" specializes in embodied control.

\begin{figure}[h]
    \centering
    \includegraphics[width=0.35\textwidth,clip,trim={5 5 5 5}]{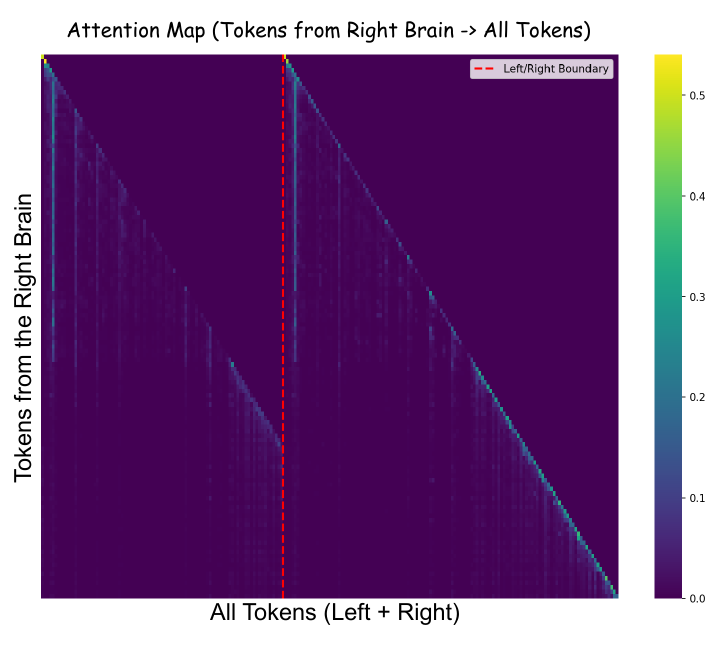}
    \caption{\textbf{TwinBrainVLA Attention Map Visualization at Runtime.} It can be seen that the right brain attends to the semantic information of the left brain via the AsyMoT mechanism.}
    \label{fig:attn_map}
    \vspace{-15pt}
\end{figure}

\subsection{The Necessity of General Semantics for Embodied Intelligence}

The fundamental premise of VLA models is to harness the extensive general capabilities acquired during the pre-training of VLMs to achieve generalized embodied intelligence. However, current VLA training paradigms often suffer from severe overfitting to specific robotic datasets~\citep{limited_diversity}, leading to "catastrophic forgetting" of the backbone's original capabilities. Addressing this paradox is the primary motivation behind \methodname{}. \methodname{} addresses this by explicitly preserving the backbone's semantic integrity. Our results confirm that retaining general capabilities directly boosts performance: our approach matches commercial models (e.g., $\pi_{0.5}$) without massive pre-training. Furthermore, this preserved knowledge enables robust instruction following and semantic generalization in real-world scenarios, handling complex tasks that traditional fine-tuning struggles to address."



\subsection{"Twin-to-One" via Knowledge Distillation}

While TwinBrainVLA successfully achieves the utilization of general capabilities by incorporating a frozen VLM as a semantic anchor, we acknowledge that this dual-stream architecture inevitably introduces additional computational overhead during deployment. To bridge the gap between high-level capability utilization and low-latency deployment, we conducted a preliminary exploration into a "Twin-to-One" knowledge distillation paradigm. In this experiment, we employed the feature representations from the Right Brain as a source of supervisory signals to re-train a Vanilla VLA architecture. The results revealed a tangible improvement in performance, which substantiates that TwinBrainVLA achieves superior training outcomes through the effective fusion of the Left Brain. This finding also highlights the potential for deeper exploration into internalizing these fused capabilities into simpler architectures. The details of these experiments are provided in Appendix~\ref{app:twin-to-one}.

\section{Conclusion}
\label{sec:conclusion}


In this work, we identified a critical paradox in current VLA models: the fine-tuning process required for robotic control inevitably leads to the "catastrophic forgetting" of the VLM backbone's general capabilities. In order to harness the pre-trained general capabilities of VLMs to achieve generalization in embodied AI, we introduce \textbf{\methodname{}\methodlogo{}} which is a framework that structurally decouples multimodal understanding from sensorimotor control. By orchestrating a frozen "Left Brain" for open-world understanding and a trainable "Right Brain" for embodied action via Asymmetric Mixture-of-Transformers (AsyMoT) mechanism, our architecture allows the policy to query high-level semantics without corrupting them. Extensive empirical evaluations on SimplerEnv, RoboCasa benchmarks and real-robot experiments demonstrate that \methodname{} significantly outperforms mainstream VLAs.

While \methodname{} marks a meaningful step towards general embodied AI, the challenges of achieving higher efficiency and deeper semantic fusion remain. We provide a detailed discussion on these limitations and future directions in Appendix~\ref{app:limitation}.

\bibliography{custom}

\newpage

\appendix
\section{Real Robot Experiments Detail}
\label{app:real-robot}

This section describes the training implementation and evaluation setup for our real-robot experiments.

\subsection{Real Robot Pre-training}

Table~\ref{tab:oxe_subsets} presents the OXE subsets used for our pre-training.

\begin{table}[h]
    \centering
    \small
    \caption{\textbf{Selected OXE Subsets for Pre-training.} We curate 20 subsets from the Open X-Embodiment dataset that utilize end-effector position control. The table details the dataset source, the robot platform, and the number of episodes utilized.}
    \label{tab:oxe_subsets}
    \setlength{\tabcolsep}{15pt} 
    \begin{tabular}{llc}
        \toprule
        \textbf{Dataset Name} & \textbf{Robot} & \textbf{Episodes} \\
        \midrule
        Berkeley Bridge & WidowX & 25,460  \\ 
        RT-1 Robot Action & Google Robot & 79,499 \\
        DROID & Franka & 92,233 \\
        BC-Z & Google Robot & 39,350 \\
        Stanford HYDRA & Franka & 550 \\
        Austin VIOLA & Franka & 135 \\
        Berkeley Autolab UR5 & UR5 & 896 \\
        NYU Franka Play & Franka & 456 \\
        Stanford Kuka Multimodal & Kuka iiwa & 3,000 \\
        Freiburg Franka Play & Franka & 3,242 \\
        USC Jaco Play & Jaco 2 & 976 \\
        NYU VINN & Hello Stretch & 435 \\
        Austin BUDS & Franka & 50 \\
        UCSD Kitchen & xArm & 150 \\
        CMU Franka Pick-Insert Data & Franka & 520 \\
        Austin Mutex & Franka & 1,500 \\
        Berkeley Fanuc Manipulation & Fanuc Mate & 415 \\
        CMU Play Fusion & Franka & 576 \\
        CMU Stretch & Hello Stretch & 135 \\
        DobbE & Hello Stretch & 5,208 \\
        \midrule
        \textbf{Total} &  & \textbf{254,786} \\
        \bottomrule
    \end{tabular}
    \vspace{-2em}
\end{table}\textbf{}

\subsection{Data Collection via Teleoperation}

\begin{figure}[htbp]
    \centering
    \includegraphics[width=1.0\textwidth]{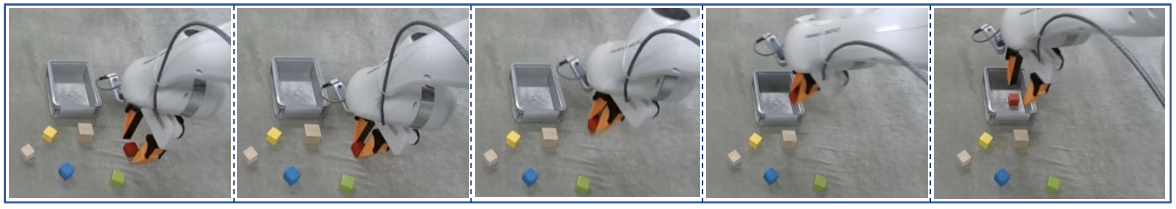}
    \caption{Diagram of human expert data acquired through teleoperation}
    \label{fig:human_expert_data}
\end{figure}

We use Polymetis~\citep{polymetis} for teleoperation and collect human expert data to support imitation learning for VLA models. For real-world experiments, we utilize a Franka Research 3 robot arm equipped with two Intel RealSense D435 cameras. We collected 300 expert human demonstrations, following the instruction format: \textit{``pick up the \{color\} block and put it into the gray box''}. The dataset is balanced, containing 100 trajectories for each of the three colors: red, green, and blue. A schematic illustration of the demonstration samples is presented in Figure~\ref{fig:human_expert_data}.

\subsection{Fine Tuning on Human Expert Demonstration Samples}

We perform fine-tuning on the checkpoints pretrained on the OXE subset, using the human expert demonstration data we collected. The training is conducted on 8 $\times$ H100 GPUs with a per-device batch size of 16 for 80K steps.

\subsection{Real Robot Evaluation}

Our evaluation is divided into three settings: in-domain, out-of-domain and pick-all.

\begin{itemize}
    \item \textbf{In-Domain}. In this setting, the evaluation tasks use exactly the same objects and task instructions as those in the training set, with the positions of the blocks altered randomly.
    \item \textbf{Out-of-Domain}. In this setting, the instructions for the evaluation tasks differ from those in the training set, primarily in terms of the color of the blocks to be grasped being altered. Specifically, the OOD evaluation task is defined by the instruction: \textit{``pick up the yellow block and put it into the gray box''}.
    \item \textbf{Pick-All}. In this setting, the task instruction is distinct from those in the training set, requiring the robotic arm to place all blocks into the box. Specifically, the task is defined by the instruction: \textit{``pick up all blocks and put them into the gray box``}, with four blocks (red, blue, green, and yellow) present in the workspace. The model is required to comprehend the task semantics and achieve long-horizon generalization by composing atomic actions learned from the training dataset.
\end{itemize}

\section{Implementation Details}
\label{app:hyperparameters}

We instantiate the VLM backbones by initializing the weights from Qwen2.5-VL-3B-Instruct and Qwen3-VL-4B-Instruct. The model is fine-tuned for 40K steps on a cluster of 16 GPUs (batch size 16 per device). We employ the AdamW~\citep{adamw_2017} optimizer initialized with a learning rate of 1e-5 and a cosine annealing schedule. System-level optimizations include DeepSpeed ZeRO-2~\citep{adamw_2017}, gradient clipping at a norm of 1.0, and no gradient accumulation. Our training pipeline is built upon the starVLA~\citep{starvla_2025} framework.

For fine-tuning, we set the default training duration to 40K steps. This choice aligns with standard practices in related work and ensures that the training can be completed within one day on NVIDIA H100 GPUs.

Our model architecture incorporates a state encoder designed to process proprioceptive information, such as joint angles and gripper states. However, since mainstream simulation benchmarks (e.g., SimplerEnv) primarily evaluate vision-only policies and do not typically involve state observations in their standard protocols, we omitted the state encoder during all simulation experiments to ensure fair comparison. Conversely, in real-robot settings where precise proprioception is crucial for control, the state encoder was active and utilized throughout both the pre-training and fine-tuning phases.

\begin{figure}[htbp]
    \centering
    \includegraphics[width=0.8\textwidth]{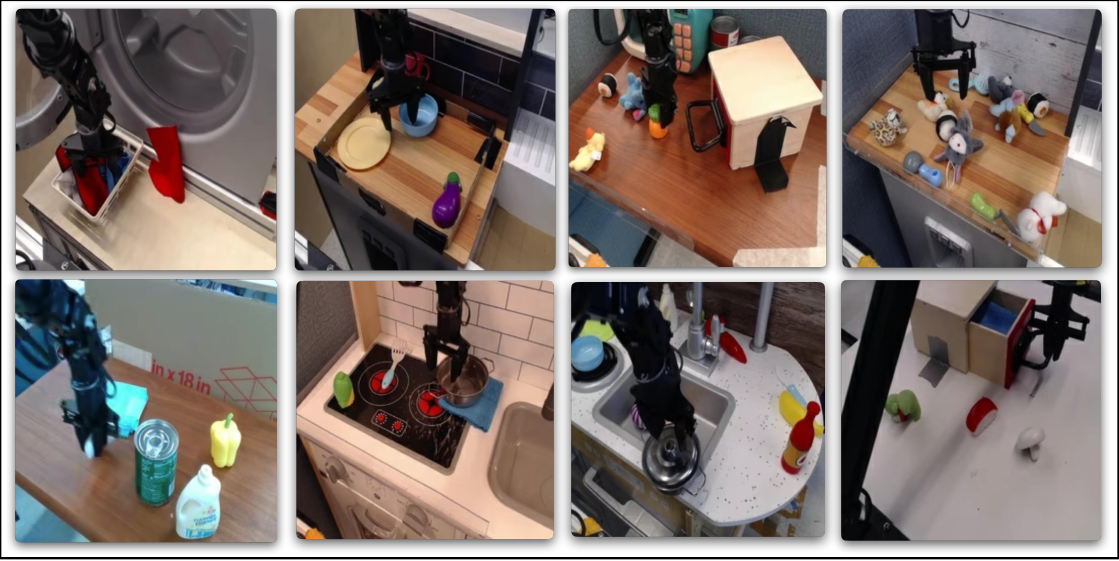}
    \caption{Examples from the Bridge V2 dataset in the training set.}
    \label{fig:oxe_visualize}
\end{figure}

\section{QwenGR00T, QwenPI and QwenOFT}
\label{app:qwengr00t}

\begin{figure}[htbp]
    \centering
    \includegraphics[width=1.0\textwidth]{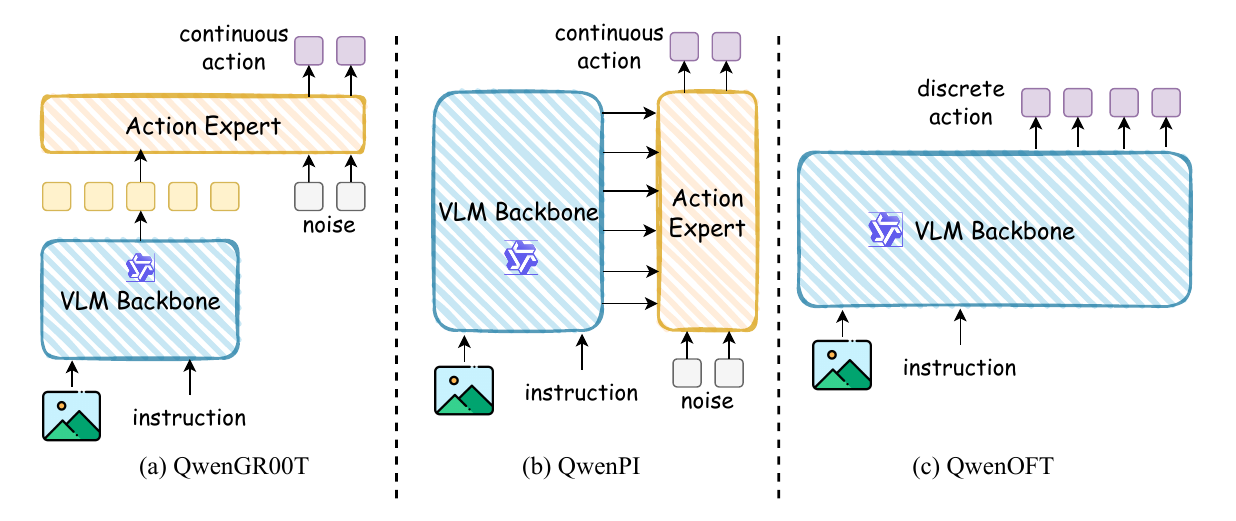}
    \caption{\textbf{Architectural overview of VLA baselines with distinct fusion strategies.} 
(a) \textbf{QwenGR00T} adopts a \textit{late-fusion} design where a flow-matching Action Expert is conditioned solely on the last hidden states of the VLM. 
(b) \textbf{QwenPI} employs a \textit{deep-fusion} strategy, enabling the flow-matching policy to interact with the VLM via layer-wise attention mechanisms. 
(c) \textbf{QwenOFT} follows the discrete token paradigm, predicting action bins autoregressively.}
    \label{fig:qwengr00t}
\end{figure}

For certain simulation experiments, we adopted the built-in baselines provided by StarVLA, re-implementing, training, and evaluating them under our uniform experimental setup. The architectures of these baselines are detailed in this section. As illustrated in Figure~\ref{fig:qwengr00t}, these models differ primarily in their action modeling and VLM-policy interaction mechanisms:

\begin{itemize}
    \item \textbf{QwenGR00T (Figure~\ref{fig:qwengr00t}a):} Incorporating a Flow Matching-based Action Expert, this model follows a \textit{late-fusion} strategy~\citep{GR00T_25}. The VLM functions as a high-level semantic encoder, passing only its last hidden states to the Action Expert. This design treats the VLM representation as a fixed holistic condition for the flow generation process.
    
    \item \textbf{QwenPI (Figure~\ref{fig:qwengr00t}b):} Also leveraging Flow Matching for continuous control, QwenPI adopts a \textit{deep-fusion} architecture~\citep{PI0}. Unlike GR00T, it establishes dense connections between the VLM and the Action Expert, allowing for layer-wise attention interactions. This enables the policy to access and leverage multi-scale semantic features from the VLM's intermediate layers during action generation.
    
    \item \textbf{QwenOFT (Figure~\ref{fig:qwengr00t}c):} Representing the discrete control paradigm~\citep{OpenVLA_24}, this model discretizes the action space into token bins and predicts actions autoregressively, treating manipulation as a standard next-token prediction task.
\end{itemize}

The implementations of the aforementioned architectures are all adopted from the built-in modules of the starVLA~\citep{starvla_2025} framework and are \textbf{not} claimed as our original contributions.

\section{RoboCasa Tabletop Benchmark}

The RoboCasa Tabletop benchmark comprises 24 distinct sub-tasks. We present the detailed success rates for all these tasks in Table~\ref{tab:robocasa_main_tab}.

\begin{figure}[h]
    \centering
    \includegraphics[width=1.0\textwidth]{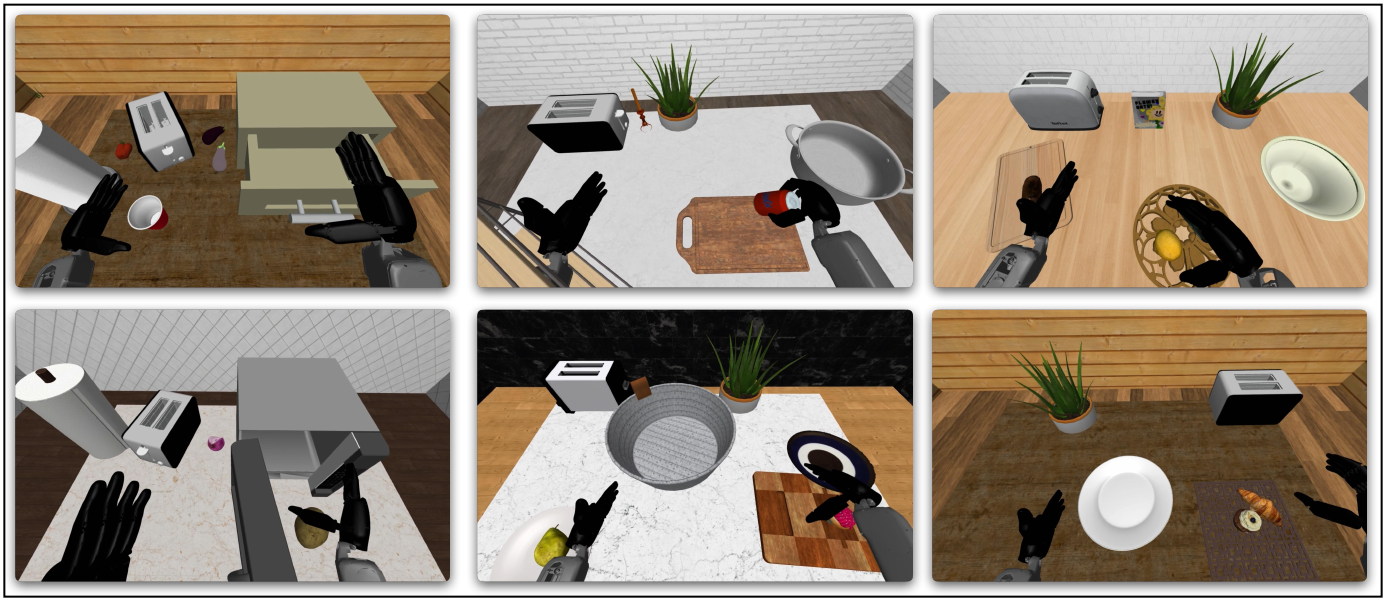}
    \caption{Schematic of RoboCasa Tabletop Evaluation.}
    \label{fig:robocasa_visualize}
\end{figure}

\begin{table*}[h]
    \centering
    \small
    \renewcommand{\arraystretch}{1.3} 
    \setlength{\tabcolsep}{5pt} 

    \caption{
      \textbf{Results of evaluating the VLA models with the GR1 robot in the RoboCasa Tabletop simulation environment}. We highlight the best results in \textbf{bold} and the second-best results with \underline{underline}.
    }
    \rowcolors{2}{white}{gray!15}

    \begin{tabular}{l c c c c c}
        \toprule
        \rowcolor{white} 
        \textbf{Task} & 
        \textbf{\scriptsize \makecell{Isaac-GR00T\\N1.6}} & 
        \textbf{\scriptsize \makecell{QwenGR00T\\ + Qwen3VL}} & 
        \textbf{\scriptsize \makecell{QwenPI\\ + Qwen3VL}} & 
        \textbf{\scriptsize \makecell{\methodname{}\\ + Qwen2.5VL}} & 
        \textbf{\scriptsize \makecell{\methodname{}\\ + Qwen3VL}} \\
        \midrule
        PnP Bottle To Cabinet Close & 51.5 & 46.0 & 26.0 & 62.0 & 74.0 \\
        PnP Can To Drawer Close & 13.0 & 80.0 & 62.0 & 66.0 & 72.0 \\
        PnP Cup To Drawer Close & 8.5 & 54.0 & 42.0 & 46.0 & 52.0 \\
        PnP Milk To Microwave Close & 14.0 & 48.0 & 50.0 & 52.0 & 60.0 \\
        PnP Potato To Microwave Close & 41.5 & 28.0 & 42.0 & 56.0 & 36.0 \\
        PnP Wine To Cabinet Close & 16.5 & 46.0 & 32.0 & 58.0 & 46.0 \\
        PnP Novel From Cuttingboard To Basket & 58.0 & 48.0 & 40.0 & 50.0 & 62.0 \\
        PnP Novel From Cuttingboard To Cardboardbox & 46.5 & 40.0 & 46.0 & 44.0 & 46.0 \\
        PnP Novel From Cuttingboard To Pan & 68.5 & 68.0 & 60.0 & 64.0 & 70.0 \\
        PnP Novel From Cuttingboard To Pot & 65.0 & 52.0 & 40.0 & 60.0 & 66.0 \\
        PnP Novel From Cuttingboard To Tieredbasket & 46.5 & 56.0 & 44.0 & 52.0 & 52.0 \\
        PnP Novel From Placemat To Basket & 58.5 & 42.0 & 44.0 & 56.0 & 30.0 \\
        PnP Novel From Placemat To Bowl & 57.5 & 44.0 & 52.0 & 59.0 & 54.0 \\
        PnP Novel From Placemat To Plate & 63.0 & 48.0 & 50.0 & 62.0 & 64.0 \\
        PnP Novel From Placemat To Tieredshelf & 28.5 & 18.0 & 28.0 & 28.0 & 38.0 \\
        PnP Novel From Plate To Bowl & 57.0 & 60.0 & 52.0 & 62.0 & 60.0 \\
        PnP Novel From Plate To Cardboardbox & 43.5 & 50.0 & 40.0 & 52.0 & 58.0 \\
        PnP Novel From Plate To Pan & 51.0 & 54.0 & 36.0 & 56.0 & 56.0 \\
        PnP Novel From Plate To Plate & 78.7 & 70.0 & 48.0 & 68.0 & 66.0 \\
        PnP Novel From Tray To Cardboardbox & 51.5 & 38.0 & 34.0 & 40.0 & 46.0 \\
        PnP Novel From Tray To Plate & 71.0 & 56.0 & 64.0 & 56.0 & 72.0 \\
        PnP Novel From Tray To Pot & 64.5 & 50.0 & 44.0 & 60.0 & 56.0 \\
        PnP Novel From Tray To Tieredbasket & 57.0 & 36.0 & 50.0 & 42.0 & 46.0 \\
        PnP Novel From Tray To Tieredshelf & 31.5 & 16.0 & 28.0 & 34.0 & 28.0 \\
        \midrule
        \rowcolor{gray!30} 
        \textbf{Average} & 47.6 & 47.8 & 43.9 & \underline{53.5} & \textbf{54.6} \\
        \bottomrule
    \end{tabular}
    \label{tab:robocasa_main_tab}
\end{table*}

\section{VLM Prompt Template}
\label{vlm_prompt_template}

This section presents the prompt templates used for VLM input during the VLA training process.

\begin{promptbox}[title={System Prompt Template for Left Brain}]
You are a helpful robot brain that can understand images and texts.
\end{promptbox}

\begin{promptbox}[title={System Prompt Template for Right Brain}]
You are a helpful robot brain that can understand images, texts, and robot states.\\
You will be provided with observation, an instruction, and the robot state. Take action to execute the instruction.
\end{promptbox}

\begin{promptbox}[title={User Prompt Template for Left Brain}]
<image>\\
Instruction: \{instruction\}
\end{promptbox}

\begin{promptbox}[title={User Prompt Template for Right Brain}]
<image>\\
Instruction: \{instruction\} \\
Robot Type: \{robot\_type\} \\
Predict the next action for the robot based on the observation, instruction and robot type.
\end{promptbox}

\section{Implementation Details on the LIBERO Benchmark}
\label{app:libero}

\textbf{Training Dataset} To evaluate the model's multi-task generalization capabilities, we adopted a unified training strategy. We aggregated the training datasets from all four LIBERO task suites (LIBERO-Spatial, -Object, -Goal, and -Long) into a single combined dataset. Instead of training separate experts for each suite, we trained a single generalist model on this unified data.

\textbf{Training Configuration.} We utilized Qwen3-VL-4B as the visual-language backbone. The model was trained on a cluster of 8 NVIDIA H100 GPUs. We set the per-device batch size to 16, resulting in a global batch size of 128. The training process was conducted for a total of 100,000 steps.

\textbf{Evaluation Protocol.} For evaluation, we strictly followed the official LIBERO evaluation protocol. We utilized the standard evaluation scripts provided by the benchmark. The performance was assessed across a total of 500 trials, consisting of 10 distinct tasks with 50 evaluation episodes per task. The final reported result represents the average success rate across these 500 trials.

\section{"Twin-to-One" via Knowledge Distillation}
\label{app:twin-to-one}

\subsection{Motivation and Methodology}
While \methodname{} achieves superior performance by leveraging a dual-stream architecture to utilize general capabilities, the inclusion of a second VLM inevitably increases computational overhead during deployment. To bridge the gap between high-level capability utilization and deployment efficiency, we propose a \textbf{"Twin-to-One"} knowledge distillation framework.

Our core hypothesis is that the "Right Brain" of \methodname{}, having dynamically fused semantic knowledge from the frozen "Left Brain" via AsyMoT, possesses a superior feature space compared to a standard single-stream VLA. Therefore, we treat the fully trained \methodname{} as the \textbf{Teacher} and a standard Vanilla VLA as the \textbf{Student}. The goal is to force the Student to internalize the Teacher's fused general capabilities by mimicking its internal representations.

\textbf{Distillation Objective.} We utilize the last hidden states of the Teacher's Right Brain as the supervision signal. Formally, let $H_{teacher} \in \mathbb{R}^{L \times D}$ denote the final hidden states of the Right Brain in \methodname{}, and $H_{student} \in \mathbb{R}^{L \times D}$ denote the final hidden states of the Student Vanilla VLA. In addition to the standard action prediction loss $\mathcal{L}_{action}$, we introduce a feature alignment loss $\mathcal{L}_{feat}$:

\begin{equation}
    \mathcal{L}_{feat} = \frac{1}{L} \sum_{i=1}^{L} \| H_{student}^{(i)} - \text{sg}(H_{teacher}^{(i)}) \|^2_2
\end{equation}

where $\text{sg}(\cdot)$ indicates the stop-gradient operation, ensuring the Teacher remains fixed. The total training objective for the Student is:

\begin{equation}
    \mathcal{L}_{total} = \mathcal{L}_{action} + \lambda \mathcal{L}_{feat}
\end{equation}

where $\lambda$ is a balancing coefficient (set to 1.0 in our experiments).

\subsection{Experiment Setup}
We conducted the distillation experiment on the SimplerEnv benchmark. The training data (Bridge-V2 and Fractal subsets) and hyperparameters (e.g., learning rate, batch size, optimizer) are kept strictly identical to the main experiments described in Appendix~\ref{app:hyperparameters}. The Student model architecture is identical to the Right Brain in TwinBrainVLA.

\subsection{Results and Analysis}
The quantitative results are presented in Table~\ref{tab:distillation_results}.

\begin{table}[h]
\centering
\caption{\textbf{"Twin-to-One" Distillation Results on SimplerEnv.} We compare the distilled single-stream model (Student) against the standard Vanilla VLA and the Teacher (\methodname{}). All models use the Qwen3-VL-4B-Instruct backbone.}
\label{tab:distillation_results}
\resizebox{1.0\linewidth}{!}{
\begin{tabular}{l|cccc|c}
\toprule
\textbf{Method} & \textbf{Spoon on Towel} & \textbf{Carrot on Plate} & \textbf{Stack Block} & \textbf{Eggplant in Basket} & \textbf{Average} \\
\midrule
\rowcolor{gray!10} \textit{Baselines} & & & & & \\
Vanilla VLA (Standard Training) & 87.5 & 50.0 & 29.2 & 54.2 & 55.2 \\
\midrule
\rowcolor{gray!10} \textit{Teacher} & & & & & \\
\textbf{\methodname{}} (dual-stream) & 87.5 & 58.3 & 33.3 & 79.1 & \textbf{64.5} \\
\midrule
\rowcolor{blue!5} \textit{"Twin-to-One"} & & & & & \\
Distilled Student (single-stream) & 83.3 & 54.2 & 29.2 & 66.7 & \underline{58.4} \\
\bottomrule
\end{tabular}
}
\end{table}

\textbf{Performance Gains.} As shown in Table~\ref{tab:distillation_results}, the Distilled Student achieves an average success rate of 58.4\%, outperforming the standard Vanilla VLA (55.2\%) by +3.2\%.

\textbf{Conclusion.} These results substantiate that the performance gains of \methodname{} stem from the high-quality, semantics-rich features fused from the Left Brain. Furthermore, it demonstrates that \methodname{} can serve as an effective supervisor, enabling single-stream models to internalize general capabilities and break the ceiling of standard fine-tuning paradigms.

\section{Limitation and Future Work}
\label{app:limitation}

\noindent\textbf{Efficiency.} The addition of a second VLM component in \methodname{} leads to higher training and deployment costs compared to vanilla VLAs. While our approach prioritizes generalizability over efficiency, we acknowledge these overheads as a limitation. Nevertheless, it is encouraging that our architecture, consisting of two 4B models, outperforms a larger 8B-parameter model, demonstrating the effectiveness of this design.

\noindent\textbf{Towards Deeper Semantic Fusion.} Although our architecture successfully enables the retention and retrieval of general capabilities, how to seamlessly fuse this semantic information into downstream VLA tasks for true generalization is not yet fully resolved. This paper identifies a viable pathway towards this goal, but achieving a definitive solution requires further exploration. We believe that more advanced mechanisms for integrating high-level semantics with low-level policies will be a critical area for future investigation.

\end{document}